\documentclass[10pt,twocolumn,letterpaper]{article}

\usepackage{cvpr}              
\usepackage{algorithm}
\usepackage{algorithmic}
\usepackage{tabularx}
\usepackage{booktabs}
\usepackage{graphicx}
\usepackage{changes}
\usepackage[most]{tcolorbox}

\definecolor{cvprblue}{rgb}{0.21,0.49,0.74}
\usepackage[pagebackref,breaklinks,colorlinks,allcolors=cvprblue]{hyperref}

\def\paperID{*****} 
\def\confName{CVPR}
\def\confYear{2026}

\title{ARIS: Agentic and Relationship Intelligence System for Social Robots}

\author{Stavya Datta\\
Monash University\\
\and
Fucai Ke\\
Monash University\\
\and
Leimin Tian\\
Monash University\\
\and
Hamid Rezatofighi\\
Monash University\\
}

\begin{document}
\maketitle

\begin{figure}[t]
    \centering
    \includegraphics[width=0.95\columnwidth]{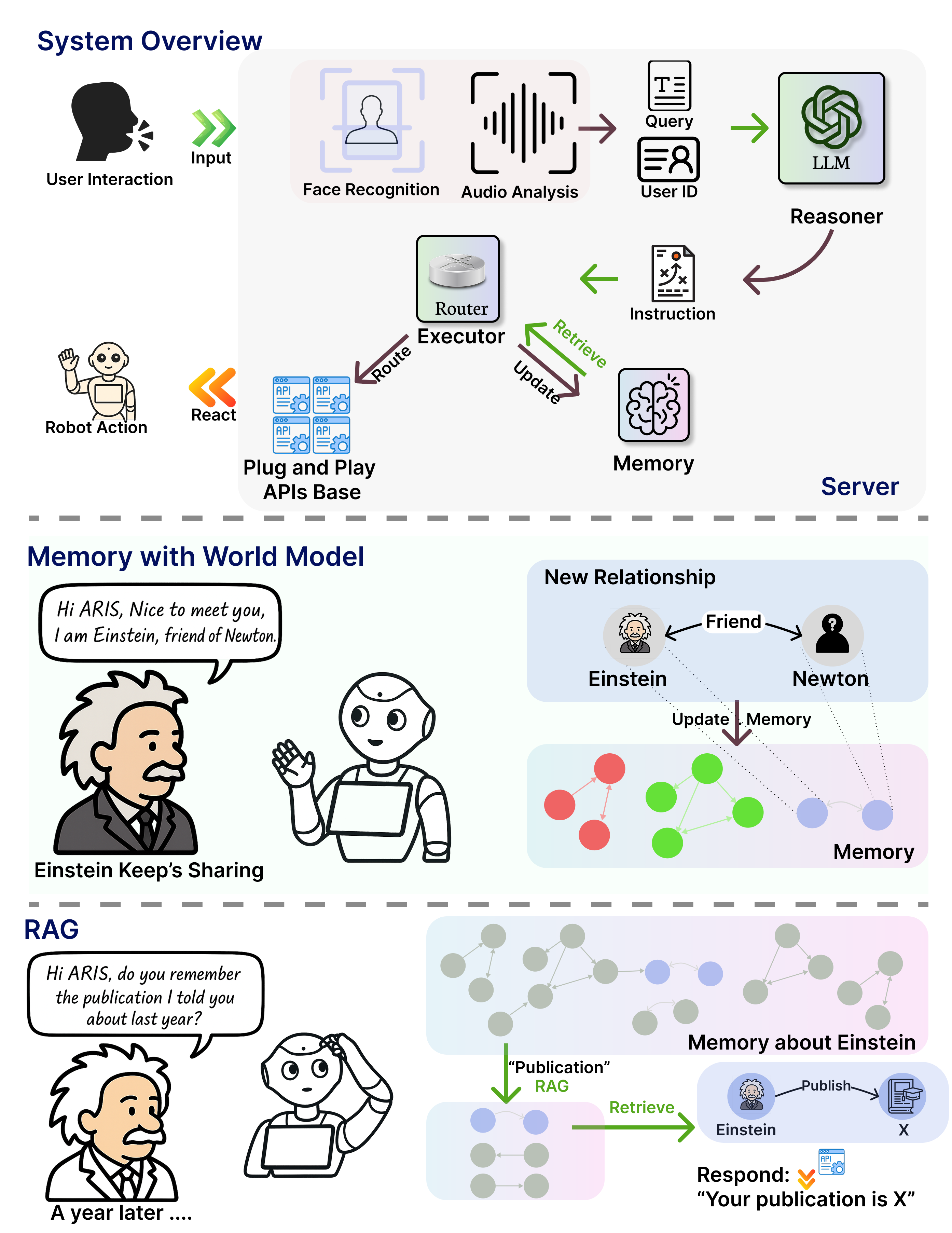}
    \vspace{-0.5em}
    \caption{Overview of ARIS, the Orchestrator for Agentic AI in Social Robots, which captures social relationships in contextually grounded multi-turn interaction via multimodal reasoning, graph-based world model, and retrieval-augmented generation.}
    \label{fig:server_diagram}
    \vspace{-1.5em}
\end{figure}

\begin{abstract}
Foundational models have advanced social robotics, enabling richer perception and communicative interaction with users. However, current systems still struggle with multi-turn engagement, social-relationship reasoning, and contextually grounded dialogue at scale. We present \textit{ARIS} (Agentic and Relationship Intelligence System), an agentic AI framework that unifies multimodal reasoning, a graph-based Social World Model, and retrieval-augmented generation (RAG) within a single modular architecture for social robots. We evaluate ARIS with the Pepper robot in a robot-mediated dyadic conversational setting, comparing it against a large language model baseline. A user study (N=23) shows that ARIS yields significantly higher perceived intelligence, animacy, anthropomorphism, and likeability. Our contributions are threefold: (1)~a Social World Model that explicitly maps and updates social relationships between users through a knowledge graph, enabling social reasoning and re-identification across encounters; (2)~an efficient RAG-based conversational pipeline that maintains bounded latency as dialogue histories grow to thousands of exchanges while preserving response relevance; and (3)~system integration and empirical validation of these components within a modular agentic architecture that coordinates speech, vision, and physical action through structured APIs. The implementation of ARIS will be released as open source upon publication.

\end{abstract}    
\section{Introduction}
\label{sec:intro}

Social robots are envisioned as agents capable of interacting with people in a natural manner, contributing to diverse applications in quality of life, entertainment, and communication \cite{breazeal_social_2016}. This vision is increasingly realized in the medical and care sectors, where robots manage social triage and elderly support tasks to alleviate administrative burdens while providing critical cognitive stimulation and reducing anxiety through persistent engagement. Several platforms have championed these impactful applications, including hyper-realistic androids like Sophia \cite{fuchs_understanding_2024}, and stylized robots such as Jibo \cite{noauthor_jibo_nodate}, paving the way for integrating social robotics into daily human life.

As these robots enter more complex real-world settings, sustaining meaningful interaction demands increasingly advanced cognitive capabilities. The emergence of large foundational models has further enhanced interactive experiences~\cite{kim_understanding_2024, schreiter2025evaluatingefficiencyengagementscripted}, with Large Language Models (LLMs) and Vision Language Models (VLMs) playing a central role in architectures such as RobotGPT \cite{chen_robogpt_2024}, SayCan \cite{ahn_as_2022}, Gemini Robotics~\cite{team_gemini_2025}, and BUMBLE \cite{shah_bumble_2024}. These models enable high-level reasoning over language, perception, and action, supporting both movement planning and natural language interaction. However, these systems primarily target task execution and physical manipulation rather than social intelligence or communicative competence. As a result, robots are limited in their ability to engage in nuanced, context-aware social dialogue.

These limitations are especially pronounced in multi-human settings, where interaction is inherently relational. Existing systems cannot represent or reason about social relationships among individuals (e.g., family members or colleagues), hindering interpretation of social roles and interpersonal dynamics. Furthermore, dense, long-term interactions exacerbate these issues: standard LLM-based architectures suffer from finite context windows and increasing inference latency, making it difficult to maintain both responsiveness and historical consistency.

To address these gaps, we propose the Agentic and Relationship Intelligence System (ARIS), a unified framework designed to imbue social robots with relational intelligence and the capacity for intense conversational depth. As shown in Figure~\ref{fig:server_diagram}, ARIS integrates a modular reasoning engine with a graph-based ``Social World Model'' to interpret social structures, alongside a context retrieval pipeline based on RAG, optimized for real-time engagement. While the architecture is designed to persist memory across distinct encounters, this work specifically validates its ability to handle the high cognitive load of extended, continuous interactions within a single session.

The contributions of this work are as follows:

\begin{enumerate}

    \item \textbf{A Social World Model for Relational Reasoning in human-robot interaction (HRI):} We introduce a graph-based representation that overcomes the limitation of flat memory systems, which cannot model or reason about human–human relationships. This enables robots to track social connections, re-identify users over time, and generate context-aware dialogue grounded in relational information, leading to more natural and socially aware interactions.
    \item \textbf{An Efficient Retrieval Pipeline for Dense Social Interaction:} We present a novel retrieval-augmented generation pipeline that efficiently manages large-scale conversational histories through hybrid semantic and recency-aware retrieval. This enables the robot to sustain contextually grounded dialogue under long-term, information-dense interactions, resulting in low-latency responses and improved relevance.
    \item \textbf{System Integration and Empirical Validation:} We unify these components within a modular agentic architecture (ARIS) that coordinates speech, vision, and physical action through structured APIs. A user study (N=23) comparing ARIS against an LLM-only baseline shows significant improvements in perceived intelligence ($d = 0.74$), animacy ($d = 0.70$), anthropomorphism ($d = 1.05$), and likeability ($d = 0.46$).
\end{enumerate}

\section{Related Work}
\subsection{Agentic AI Systems}
The proliferation of LLMs has inspired agentic AI and reasoning architectures that couple these models with external tools \cite{ke_explain_2025}. Toolformer \cite{schick_toolformer_2023} integrates Python interpreters and web search engines, while compositional reasoning works like ViperGPT \cite{suris_vipergpt_2023} and HYDRA \cite{ke_hydra_2024} focus on planning and reasoning for complex tasks. While these works have made significant strides in task-oriented reasoning, their architectures are not directly transferable to live social robot systems. Taking inspiration from these works we have developed the ARIS system, described in Section~\ref{sec:system_architecture_explanation}.

\subsection{Social Robotics with LLMs}
In social robotics, LLMs have advanced the capabilities of conversational agents. RobotGPT \cite{chen_robogpt_2024} empowers robots to comprehend and generate natural dialogue across diverse social contexts. Recent work by Yaxin et al.\ \cite{hu_this_2024} demonstrates the potential of LLM-powered robots for specific populations such as elderly bedridden patients. Similarly, Kim et al.\ \cite{kim_understanding_2024} have shown that integrating LLMs with the Pepper robot enables more naturalistic conversations in public environments. We have also seen humanoid robots with nuanced physical embodiment abilities such as Gemini Robotics \cite{team_gemini_2025} and BUMBLE \cite{shah_bumble_2024}. However, these systems prioritize task execution and physical manipulation over social intelligence and communicative competence. As noted by Breazeal et al.\ \cite{breazeal_social_2016}, effective social robots require not merely functional capabilities but explicit models of social cognition, interaction patterns and communication skills that mirror human social dynamics.

\subsection{Social World Model and RAG}
Social World Model here refers to an internal cognitive structure that allows individuals to simulate and predict outcomes based on the environment \cite{johnson-laird_mental_2010}. Nigro et al.~\cite{nigro_social_2025} underscore the importance of such representations in social robotics and HRI, identifying the detection of interpersonal relationships and subgroups as a key challenge. Garello et al.~\cite{garello_building_2025} operationalize this concept as a \emph{Knowledge Graph}, representing structured experiences of the robot, including entities such as users, activities, and interactions, to organize memories for efficient retrieval and generalization.

Retrieval-Augmented Generation (RAG) enhances LLMs by integrating external information retrieval mechanisms~\cite{gupta_comprehensive_2024}. RAG has been employed to ground LLM outputs with relevant documents \cite{gupta_comprehensive_2024, izacard_few-shot_2022} and utilised in robotics for task planning \cite{long_drae_2025, ginting_saycomply_2025} and domain-specific document retrieval \cite{fazlollahtabar_human-robot_2025}. However, to the best of our knowledge, dense interaction quality and efficiency in social robotics remain underexplored. Therefore, we propose an efficient mechanism for processing dense one-to-one conversations using a RAG approach. It is worth noting that unlike the robot-mediated group conversations surveyed by Nigro et al., our system focuses on dyadic interactions, where the robot builds an internal social map of relationships rather than mediating group discourse.

\subsection{Robotic Planning and Reasoning}
Integrating Large Foundational Models (LFMs) has shifted robotics from simple execution to complex, multi-stage task planning. Systems like BUMBLE \cite{shah_bumble_2024} use Vision-Language Models and hierarchical memory for building-scale navigation and manipulation. Simultaneously, social robotics has advanced through hybrid models based on ACT-R methodology, which enable personalized user interactions \cite{sievers_retrieving_2025}.

However, a critical gap remains: current frameworks typically treat functional tasks and social intelligence as separate domains. Existing literature lacks a unified architecture that optimizes for social intelligence. ARIS tries to bridge that gap using its Agentic AI System, Social World Model and RAG based conversation system.


\section{System Architecture}
We now describe the ARIS architecture. Section~\ref{sec:system_architecture_explanation} outlines the overall system, Section~\ref{sec:db_world_model_explanation} details the Social World Model, and Section~\ref{sec:rag_implementation} presents the RAG-based speech pipeline.

\subsection{System Architecture Overview}
\label{sec:system_architecture_explanation}
\subsubsection{System Overview}
ARIS follows a \textbf{edge-client-server architecture}, as illustrated in Figure~\ref{fig:server_diagram}. It integrates three core components: Edge, Client, and Server. 
The \emph{Edge}, represented by Pepper Robot v1.8 \cite{pandey_mass-produced_2018}, serves as the user-facing interface. The \emph{Client} governs robot control by capturing audio-visual streams, encapsulating them into gRPC packets, and transmitting them to the server. In return, the Edge interprets server responses as plain text for spoken output or as JSON commands for actions such as gestures, high-fives, or dances.  

The \emph{Server} functions as a gRPC endpoint that processes these multimodal inputs. Audio is transcribed using Whisper-large-v3 \cite{radford_robust_2022}, enabling natural speech interactions. The Reasoner interprets the transcribed input and determines suitable actions, while the Executor module executes these actions on the robot.

To support diverse robot functionalities, we propose a plug-and-play API base that enables designers to flexibly customize system capabilities according to application needs. In addition, we introduce a dedicated memory module that functions as a Social World Model knowledge graph, embedding the complete history of each user’s interactions with the robot. This representation captures user attributes, activities, and relationships with others, thereby enabling more personalized and socially aware behaviors. We further integrate a Re-Identification (ReID) module that processes image streams from the robot’s vision system; through majority voting across $N=15$ frames, it assigns each individual a \texttt{Person Node}, retrieving the corresponding face\_id when a match exists or generating a new one otherwise.

\subsubsection{Reasoner} 
\label{sec:reasoner_explanation}
The Reasoner receives the transcription and \texttt{face\_id}, querying the user's details from the database. It consults the \texttt{API} library, which documents available APIs and their functionalities, and determines which APIs to invoke based on the transcription context. Current APIs span Speech, Vision, and Physical Embodiment, and the Reasoner flexibly selects whichever combination best fits the task.

\subsubsection{Executor} 
\label{sec:executor_explanation}
The Executor routes Reasoner instructions to the appropriate API and standardizes responses for the client. Key APIs include:

\begin{itemize}
    \item \textbf{Speech API}: The API uses Large Language Model and RAG based system to respond to the input coming from the users, more information in Section~\ref{sec:rag_implementation}
    \item \textbf{Vision System}: The Vision System is activated when user input requires vision capabilities. We use a VLM as a captioner to respond to the final query.  
    \item \textbf{Physical Embodiment}: This module is executed when the user input requires specific movement like a high-five or dances.
\end{itemize}


\subsubsection{Technical Specifications}
Table~\ref{tab:system_specs} summarizes the models and tools used in each system component.

\begin{table}[ht]
\centering
\caption{System Component Specifications}
\label{tab:system_specs}
\setlength{\tabcolsep}{4pt}
\small
\begin{tabularx}{0.8\linewidth}{p{0.35\linewidth} X}
\hline
\textbf{Component} & \textbf{Model / Tool} \\
\hline
Speech Recognition & Whisper-large-v3 \\
Reasoner LLM & Grok 2 \cite{xai_grok_nodate} \\
Speech LLM & Grok 2 \cite{xai_grok_nodate} \\
Vision LM & Grok-2-Vision \\
Embedding Model & OpenAI Embeddings \cite{openai_new_2024} \\
Graph Database & Neo4j \\
Hardware (Client) & Raspberry Pi 5 \\
Hardware (Robot) & Pepper v1.8 \cite{pandey_mass-produced_2018} \\
Hardware (Server) & NVIDIA RTX 4090 \\
\hline
\end{tabularx}
\end{table}

\subsection{Social World Model} 
\label{sec:db_world_model_explanation}

\subsubsection{Social World Model Design}
The Social World Model is a graph network that stores information about users interacting with the robot, where each node corresponds to a \texttt{Person Node} representing an individual user. An edge between the \texttt{Person Nodes} represents relationship between them as shown in Figure~\ref{fig:sample-world-model}

\textbf{Person Node:} 
Each node stores key information, including the person's $name$ as inferred from the conversation, a unique $face\_id$ assigned during face recognition as discussed in Section~\ref{sec:system_architecture_explanation}, and a set of $attributes$ describing personal details such as preferences, interests, or occupation as identified by the system.

\subsubsection{Social World Model Execution}
\label{sec:world-model-desc}
\begin{figure}[tb]
    \centering
    \includegraphics[width=0.4\textwidth]{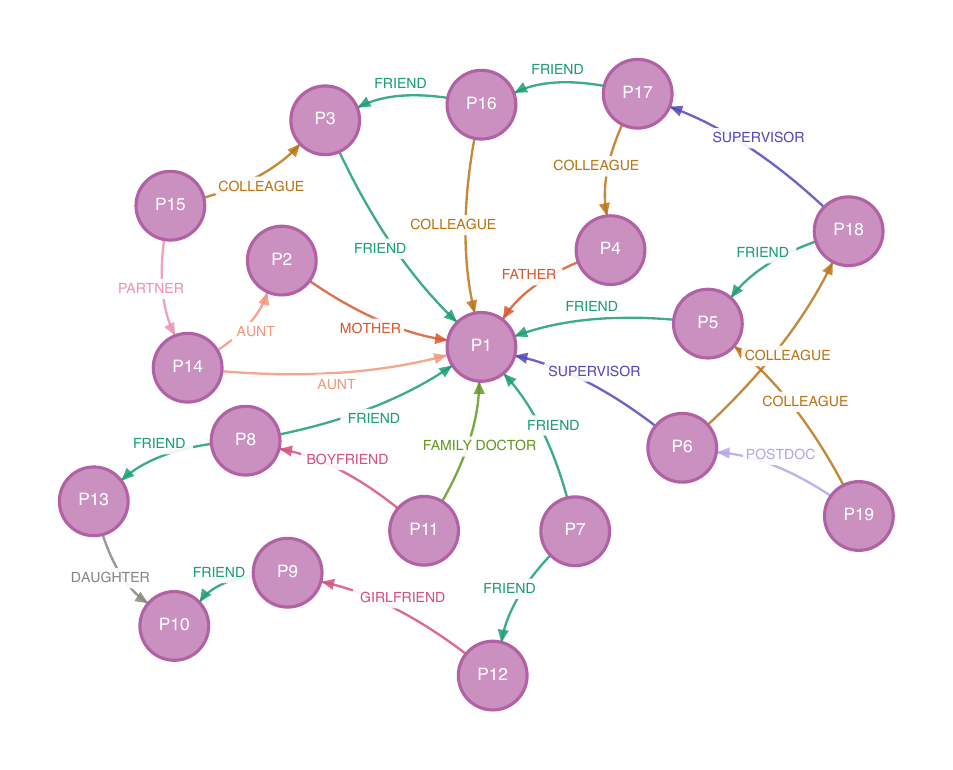}
    \caption{Example Social World Model diagram (Participants names redacted)}
    \label{fig:sample-world-model}
\end{figure}

The Social World Model enables the system to understand and represent interpersonal relationships, as illustrated in Figure~\ref{fig:sample-world-model}. During interaction, ARIS processes spoken input to infer relational contexts, such as \emph{friendships}, \emph{parent-child relationships}, and similar connections. The responsibility of relationship classification is delegated to the underlying LLM model.

Upon identifying relational cues in the user's utterance, the system extracts the relevant entity names, determining whether the relationship refers to the speaker and another individual, or between third parties. This extracted information is then formulated as a Cypher query to update the system's knowledge graph. The decision of extracting relevant details and formulation of Cypher query is assisted by LLMs

If a referenced name does not exist in the database, a new \texttt{Person Node} is created. Conversely, if the name is already present, the system simply establishes or updates the appropriate connections between entities as shown in Algorithm~\ref{algo:world_model_explanation}.

When a new user introduces themselves, the Social World Model matches the spoken name against existing nodes using the \emph{Levenshtein distance}. If a match is found, the system merges the new node with the existing \texttt{Person Node}, consolidating all related information.

\subsubsection{Attribute Development}
Analogous to relationship extraction, the system also identifies and assigns attributes to individuals. These attributes may pertain to the speaker or to a third party referenced in conversation. The determination of the attribute’s subject is handled by an LLM, which assesses whether the description concerns the speaker or another person. Attributes—such as likes, dislikes, occupation, and interests—are then stored accordingly within the relevant \texttt{Person} node.

\begin{algorithm}[h]
\caption{Relationship Development in Social World Model}
\small
\begin{algorithmic}[1]
\label{algo:world_model_explanation}
\WHILE{true}
  \STATE \textbf{msg} $\gets$ get\_user\_msg()
  \STATE \textbf{result} $\gets$ relationship\_finder\_grounded\_llm(msg)
  \IF{\,not result.\texttt{is\_relationship}\,}
    \STATE \textbf{continue}
  \ENDIF
  \STATE \textbf{userName} $\gets$ get\_user\_name()
  \STATE \textbf{thirdPersonName} $\gets$ result.\texttt{thirdPersonName}
  \STATE \textbf{names} $\gets$ userName + thirdPersonName
  \STATE \textbf{queryTemplate} $\gets$ result.\texttt{cypher\_query}
  \STATE \textbf{params} $\gets$ find\_closest\_name\_in\_db(thirdPersonName)
  \STATE \textbf{query} $\gets$ inject\_params(queryTemplate, params)
  \STATE execute\_cypher\_query(query)
\ENDWHILE
\end{algorithmic}
\end{algorithm}

\subsection{RAG based speech pipeline}
\label{sec:rag_implementation}
\subsubsection{RAG based speech pipeline design} 
The speech system in ARIS employs a RAG pipeline that retrieves contextually relevant information from prior interactions. This design addresses two key challenges: first, the potentially large volume of user history, which makes direct retrieval from raw logs computationally expensive; and second, the limited context window of large language models, which prevents efficient processing of long conversational histories.

Within this framework, the \texttt{Message Entity} (or \texttt{Message Node}) serves as the fundamental unit of conversational information, analogous to the notion of a \emph{document} in RAG literature~\cite{gupta_comprehensive_2024}. 

Each \texttt{Message Node} encapsulates a user input and the robot’s response, together with a unique $message\_id$ and an associated $face\_id$ that links the message to the relevant \texttt{Person Node}, as illustrated in Figure~\ref{fig:sample-message-graph}.  

All message texts are encoded using an embedding model (e.g., OpenAI Embeddings \cite{openai_new_2024}), producing 1536-dimensional vectors that are indexed in the Neo4j Vector Database. The database employs approximate nearest neighbor (ANN) search with cosine distance and Hierarchical Navigable Small World (HNSW) indexing, enabling rapid retrieval of semantically similar messages across large volumes of conversational data.  

\subsubsection{RAG based speech pipeline execution} 
The pipeline is executed through the \emph{Speech API}, introduced earlier in Section~\ref{sec:executor_explanation}. When a new user input is received, the API queries the database using the corresponding $face\_id$ to locate the relevant \texttt{Person Node}. From this point onward, both the user’s input and the system’s reply are referred to as \emph{messages}. If available, the person’s $attributes$ and $name$ are incorporated into the $system\_prompt$. The user input is then embedded using the same embedding model, and the resulting vector is employed to search the database. This process retrieves two sets of \emph{messages}:

\begin{itemize}
    \item The 20 closest \emph{messages }to the current input using cosine distance as metric and ANN for searching, with each accompanied by its immediate predecessor and successor to provide better context for the \emph{messages}, yielding 60 \emph{messages }in total.  
    \item The 20 most recent \emph{messages} preceding the input, providing continuity of the current dialogue.  
\end{itemize}  

The retrieved messages are consolidated into a chronological chain, combined with the most recent user input, and pruned to ensure uniqueness. This chain, capped at a maximum of 80 messages, is then provided to the LLM for response generation. After the model produces a reply, the new message is appended to the chain, and a \texttt{MESSAGE} relationship of the associated \texttt{Person Node} is updated.  

\begin{figure}[tb]
    \centering
    \includegraphics[width=0.35\textwidth]{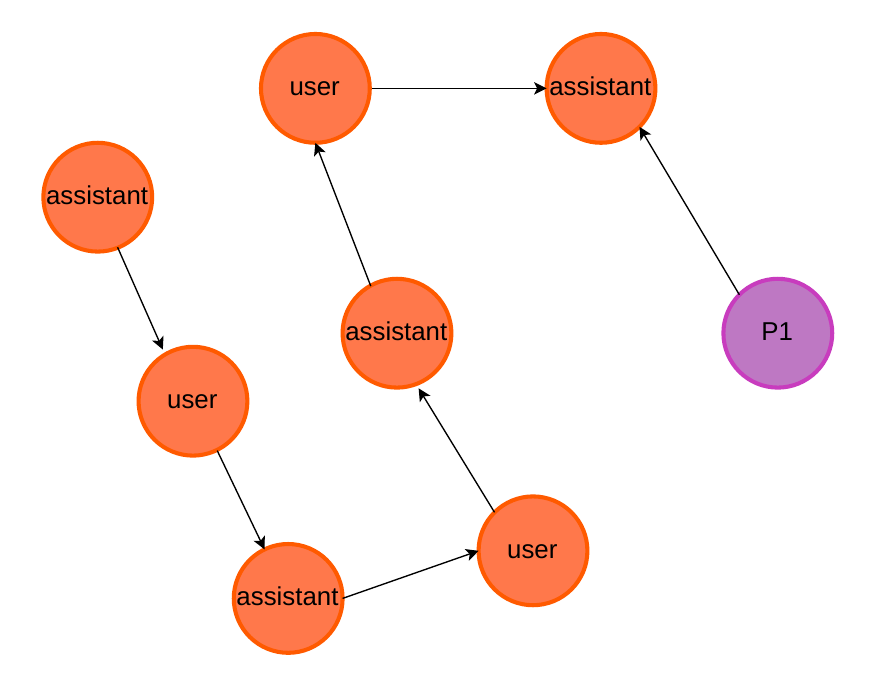}
    \caption{An example Person and Message Node, the violet one is \texttt{Person Node} and orange ones are \texttt{Message Node}. The \texttt{Message Nodes} are in an ordered chain. The latest message node is connected to the \texttt{Person Node}}
    \label{fig:sample-message-graph}
\end{figure}

This selective retrieval ensures that only semantically relevant history is supplied to the LLM, reducing computational overhead while maintaining contextual grounding. Figure~\ref{fig:sample-message-graph} illustrates how \texttt{Message Nodes} attach to their corresponding \texttt{Person Node}. The full pipeline is detailed in Algorithm~\ref{algo:speech_gen}.

\begin{algorithm}[h]
\caption{Speech Generation for Person}
\small
\begin{algorithmic}[1]
\label{algo:speech_gen}
\REQUIRE inputMessage, personObj
    \STATE faceID, personName, personAtrributes $\gets$ getPersonDetails() 
    \STATE closestMessages $\gets$ GraphDB.getClosestMessages(faceID, inputMessage)
    \STATE last20Messages $\gets$ GraphDB.getLastMessages(faceID)
    \STATE totalMessages $\gets$ unique(closestMessages + last20Messages)
    \STATE llmResponse $\gets$ SpeechLLM(totalMessages)
    \STATE YIELD llmResponse
    \STATE personObj.addMessages(inputMessage, llmResponse)
    \STATE RelationshipChecker(personObj) 
\end{algorithmic}
\end{algorithm}

\section{Experiment}
\subsection{Technical Implementation}
To evaluate the feasibility and real-time performance of our system in a realistic deployment setting, we conducted experiments with the client running on a Raspberry Pi 5 and the server running on an NVIDIA RTX 4090 GPU. We employed Pepper Robot Version 1.8. The database for the Social World Model and the RAG-based speech system was implemented with Neo4j Graph Database. All system component specifications are detailed in Table~\ref{tab:system_specs}.

\subsection{Hypotheses}
\label{sec:experiment_hypothesis}

The aim of our experiments was to systematically evaluate whether the proposed enhancements to the robot platform would measurably improve user experience and system performance. To this end, we formulated the following hypotheses:

\begin{description}
    \item[H1: Robot Perception:] ARIS is expected to yield more positive user perception of the robot compared to a baseline LLM-only system.

    \item[H2: RAG-based Scalability:] The RAG mechanism will maintain bounded inference time as conversation length increases, while producing outputs that remain comparable to a Non-RAG pipeline that has access to the full conversation history.
\end{description}

We conducted two experiments: (1) a \textbf{User Study}, testing \textbf{H1}, and (2) a \textbf{RAG vs. Non-RAG experiment}, testing \textbf{H2} (see Section~\ref{sec:rag_experiment}).

Rather than isolating individual components, our evaluation adopts a system-level comparison to assess whether a unified agentic architecture that integrates social reasoning, multimodal perception, context retrieval, and embodied action yields measurable improvements over a representative LLM-only conversational baseline. This design reflects a realistic deployment scenario in which users interact with the complete system, and mirrors the baseline reported by Kim et al.~\cite{kim_understanding_2024}.

\subsection{User Study Design}
\label{sec:only-llm-design}
\subsubsection{LLM-Only Baseline}
This system differs significantly on the \emph{server-side} from the ARIS system described in Section~\ref{sec:system_architecture_explanation}. The server pipeline consists of Whisper \cite{radford_robust_2022} for automatic speech recognition (ASR), followed by a LLM (Grok 2 in our case \cite{xai_grok_nodate}) that generates a natural language response. This response is then transmitted to the client and vocalized by the robot. Notably, the system lacks mechanisms for user re-identification, does not incorporate a Social World Model for understanding relationships between people. Furthermore, it does not include a flexible API selection or reasoning module to determine appropriate actions. As a result, the system cannot perform embodied physical tasks, nor can it adaptively decide when to initiate or withhold responses. This system is similar to what has been implemented on Pepper with Kim et al \cite{kim_understanding_2024} and will be used to compare against our ARIS system. 

\begin{figure}
    \centering
    \includegraphics[width=0.5\textwidth]{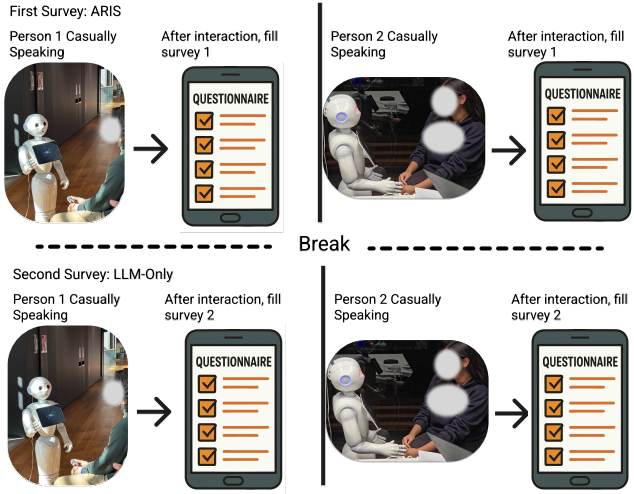}
    \caption{User Study Interaction Example}
    \label{fig:user_study_flowchart}
\end{figure}

\subsubsection{User Study Protocols}
Participants took part in our HRI experiment with each session involving two participants working in pairs. 

As depicted in Figure~\ref{fig:user_study_flowchart}, in each round, the first participant began by introducing themselves and engaging in casual conversation with the robot, sharing details about their day and hobbies. They then described their partner by stating their name, hobbies, and relationship (e.g., friend or family member). The second participant then followed the same sequence to interact with the robot. During the individual introductions, both participants are in the same room with the robot, but only one participant is interacting with the robot at a given time while the other participant observed the interaction. 

Each participant pair completed two experimental rounds. In one round, participants interacted with the baseline system, which relied exclusively on the LLM as detailed in Section~\ref{sec:only-llm-design}. In the alternate round, they engaged with the full ARIS system, incorporating speech-to-text, face re-identification, a Social World Model, physical embodiment, and an API-driven reasoning framework, as outlined in Section~\ref{sec:system_architecture_explanation} and Figure~\ref{fig:user_study_flowchart}.

A short break was provided between rounds. Participant order within each pair was held constant, participants rated each round independently without explanation of system differences, and system order was counterbalanced across pairs to mitigate order effects. Each round concluded with a questionnaire, and the entire session lasted approximately 25 to 30 minutes.

\subsubsection{Questionnaire Design}
The questionnaire was based on the Godspeed questionnaire \cite{bartneck_measurement_2009}, including semantic differential scales measuring animacy, likeability, anthropomorphism, and intelligence on a five-point Likert scale from 1 (Strongly Disagree) to 5 (Strongly Agree). The perceived safety dimension was excluded as all interactions were seated conversational tasks. Additional items examined the Social World Model’s performance (e.g., whether the robot remembered information about a participant’s partner) and the robot’s physical embodiment, including whether its movements and vision capabilities affected the interaction.

\subsection{RAG vs Non-RAG Experiment Design}
\label{sec:rag_experiment}
\subsubsection{Non-RAG Pipeline} In the Speech API, the Non-RAG setting retrieves the entire conversational history and sends it to the LLM, rather than selecting relevant messages. To test this, we stored messages as JSON in the database for each user ID, alongside the message node pipeline described in Section~\ref{sec:rag_implementation}, within the \texttt{Person Node} as an attribute. This setup emulated a real-world non-RAG pipeline in a robot comparable to \cite{kim_understanding_2024}
\subsubsection{Simulated experiment implementation}
The goal of this experiment was to evaluate whether our RAG pipeline achieves faster inference while maintaining output quality comparable to a Non-RAG baseline. 

To simulate dense conversational contexts, we selected 11 user IDs from our study and duplicated their stored messages to extend the conversational history to thousands of entries. For the Non-RAG condition, the entire history was stored in JSON format attached to each \texttt{Person Node}, emulating a system that retrieves and forwards all prior interactions to the LLM. For the RAG condition, only relevant messages were retrieved through the pipeline described in Section~\ref{sec:rag_implementation}.

We compared both pipelines on two dimensions: (1) retrieval and inference time, and (2) output similarity. For evaluation, we posed 10 personal questions per user ID and repeated them three times to measure consistency. Output similarity between RAG and Non-RAG responses was computed using the Gemma-3 embedding model 300M from Google DeepMind \cite{schechter_vera_embeddinggemma_2025}, using cosine similarity.

\subsection{Ethics Consideration}
The experiment protocols were reviewed and approved by our university's ethics board. All sessions were conducted in a designated meeting room. Participants were recruited via advertised sign-up links, provided informed consent before the session began, and were informed they could withdraw at any time without consequence. All data collected during the sessions remain confidential and stored securely on an encrypted hard drive.

\section{Results}
Our user study results are summarized in Table~\ref{tab:stat_results}. Section~\ref{sec:result_declaration} presents the key trends observed during participant interactions with both the proposed ARIS system and the LLM-only baseline. Section~\ref{sec:rag_experiment_result} reports the results of the RAG vs.\ Non-RAG pipeline experiment. Our findings demonstrate that ARIS significantly outperformed the LLM-only baseline across all perception metrics, with particularly large effects in Anthropomorphism (Cohen's $d = 1.05$) and medium effects in Intelligence and Animacy, while the RAG pipeline maintained bounded inference time as conversation length increased.

\begin{table*}[ht]
\centering
\caption{Statistical test results summary across 23 participants. Alongside $p$-values, Cohen's $d$ indicates effect magnitude.}
\label{tab:stat_results}
\begin{tabular}{lcccc}
\toprule
Metric & ARIS Mean \textpm std & LLM-Only Mean \textpm std & $t$-p & Cohen's $d$ \\
\midrule
Animacy          & $\mathbf{4.00 \pm 0.72}$ & $3.20 \pm 0.97$ & 0.0030 & 0.70 (medium) \\
Likeability      & $\mathbf{4.35 \pm 0.42}$ & $3.93 \pm 0.87$ & 0.0398 & 0.46 (small--medium) \\
Intelligence     & $\mathbf{4.15 \pm 0.73}$ & $3.35 \pm 1.0$ & 0.0018 & 0.74 (medium) \\
Anthropomorphism & $\mathbf{3.98 \pm 0.80}$ & $2.65 \pm 0.93$ & $<0.001$ & 1.05 (large) \\
\bottomrule
\end{tabular}
\end{table*}

\begin{figure}[tb]
    \centering
    \includegraphics[width=0.45\textwidth]{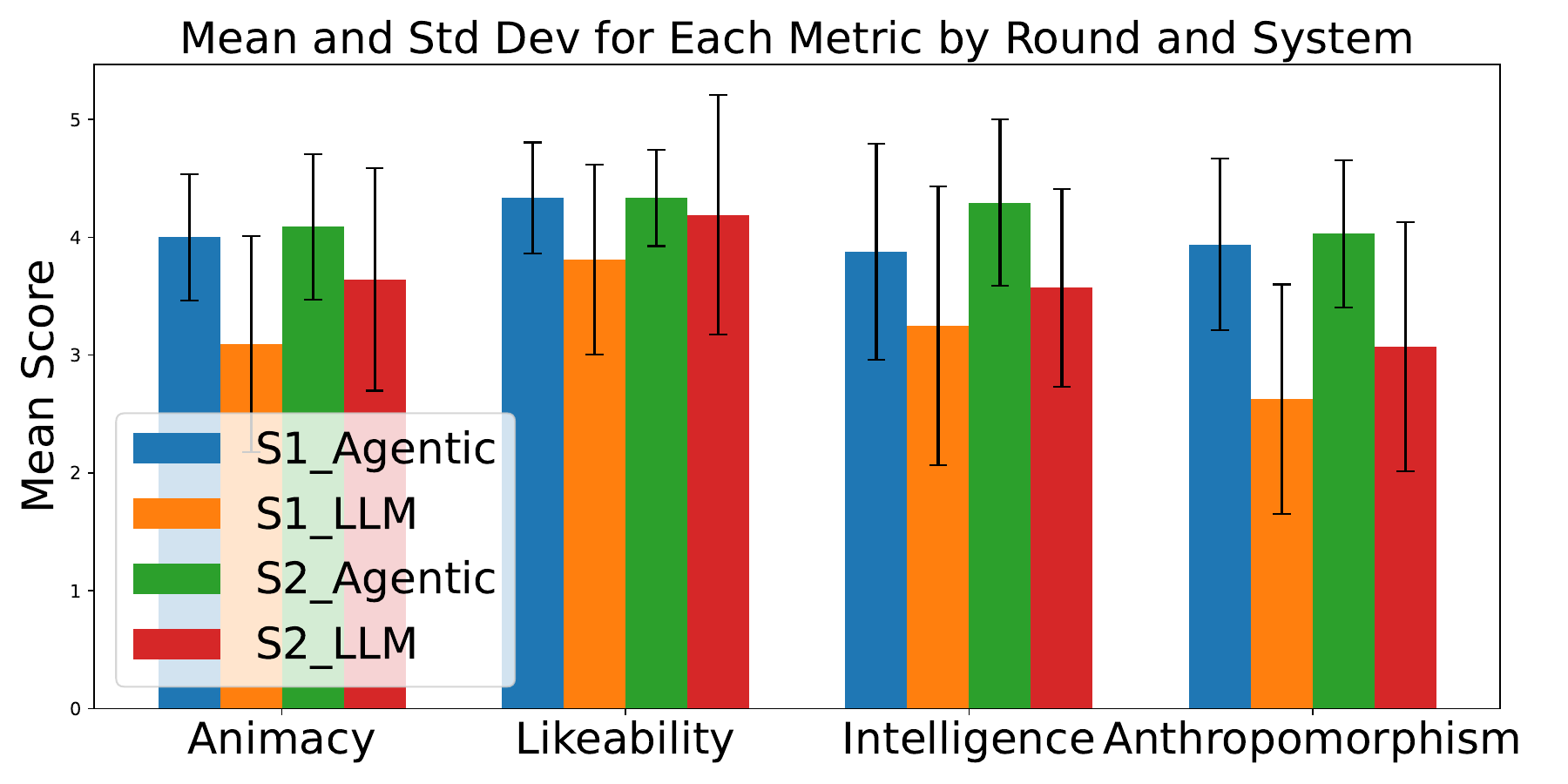}
    \caption{GodSpeed questionnaire ratings across rounds and systems. S1/S2 indicate round order, not system order. ARIS consistently outperforms the LLM-only baseline across both rounds.}
    \label{fig:Agentic_vs_LLM}
\end{figure}

\subsection{Participants}
We recruited 24 participants, organised into 12 pairs, through the university mailing list. The sample comprised 18 males and 6 females, aged between 18 and 30 years. Of these, 7 participants were aged 18–20 and 17 were aged 21–30. During the study, one male participant aged 21-30 completed the forms incorrectly by selecting the ``Agentic AI based'' (i.e ARIS) system for both surveys. Consequently, this participant's data was excluded from the final analysis.

\subsection{User Study Results}
\label{sec:result_declaration}
\subsubsection{System Performance}
As shown in Table~\ref{tab:stat_results}, ARIS achieved consistently higher mean scores across all measured constructs. Paired t-tests indicated statistically significant differences ($p<0.05$) for all four metrics: Anthropomorphism ($p < 0.001$, $d = 1.05$), Intelligence ($p = 0.0018$, $d = 0.74$), Animacy ($p = 0.003$, $d = 0.70$), and Likeability ($p = 0.0398$, $d = 0.46$). Figure~\ref{fig:Agentic_vs_LLM} shows the robot perception ratings for both ARIS and LLM-only systems, with interaction order showing no significant influence. Participants also noticed ARIS's movement capabilities, rating engagement with robot movements at 3.91 (vs.\ 2.0 for LLM-Only) and expected movement execution at 3.86 (vs.\ 2.10).

\subsubsection{Social World Model}
Beyond godspeed metrics, the Social World Model’s performance was tested using pairs of participants, where one described their partner before the other interacted with the robot. Under ARIS, the system successfully recognized 7 of 12 Second to Interact participants, with 9 of 12 reporting positively that the robot ``knew about my partner when asked.'' In contrast, under the Only-LLM baseline, 4 of 12 participants felt the robot identified them based on their partner’s description, and 4 of 12 felt it could describe their partner.

Our user study results confirm the positive impact of ARIS on user's robot perception, \textbf{validating H1}.

\subsection{RAG vs Non-RAG Simulation Results}
\label{sec:rag_experiment_result}

\begin{figure}[tb]
    \centering
    \includegraphics[width=0.45\textwidth]{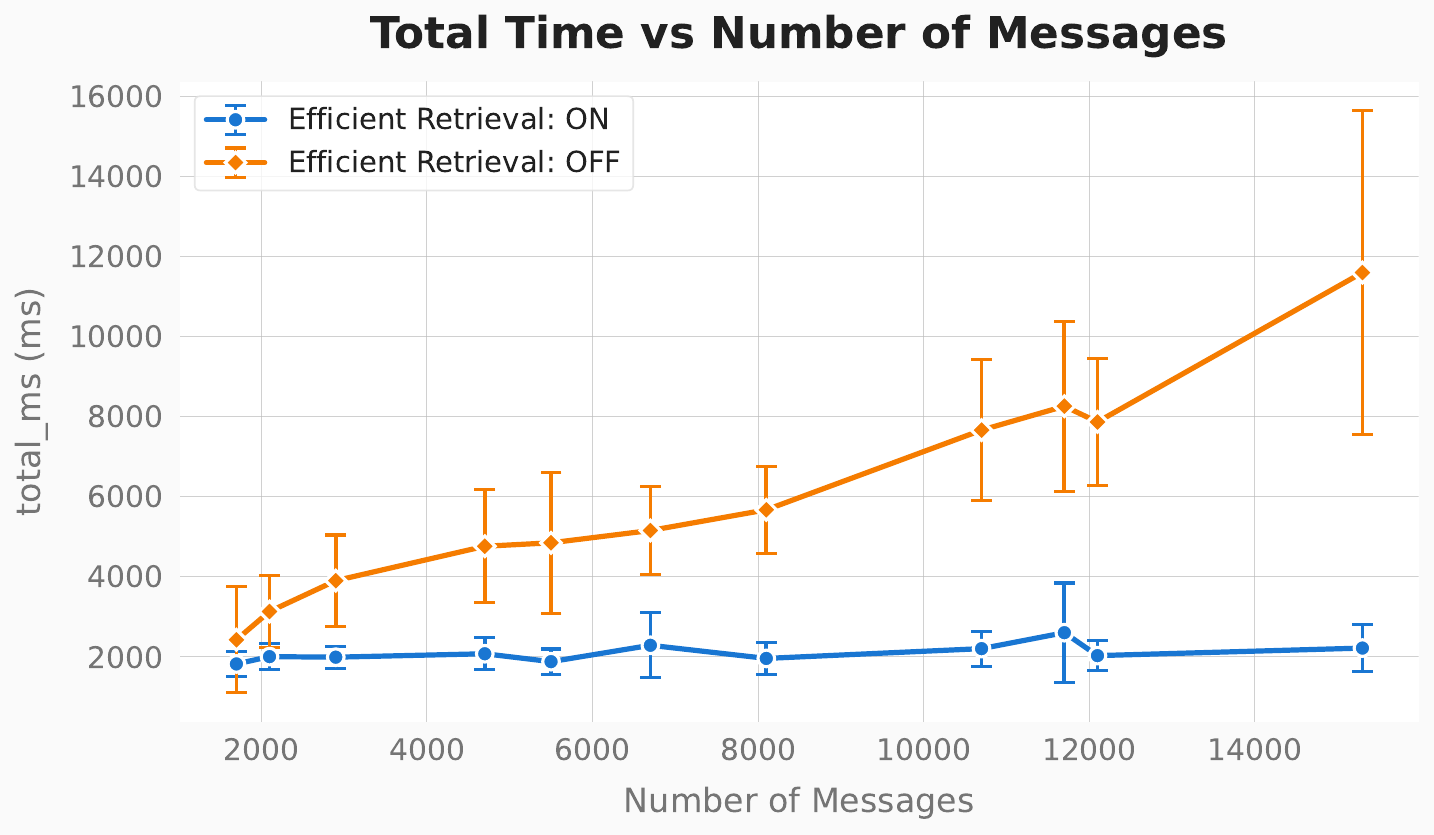}
    \caption{Comparing Retrieval + LLM Response time with number of Messages, as the Message Number increases.}
    \label{fig:rag_total_message_len}
\end{figure}
\textbf{Scalability Results.} Figure~\ref{fig:rag_total_message_len} shows retrieval plus inference time as a function of message count. The Non-RAG pipeline exhibits approximately linear-to-superlinear growth, exceeding $10,000$ ms at $14,000$ messages, whereas the RAG pipeline remains consistently below $4,000$ ms regardless of conversation length. This confirms that selective retrieval via the RAG pipeline provides bounded latency suitable for real-time interaction, even at conversation scales far exceeding typical single-session exchanges.

\textbf{Output Comparison.} As a secondary analysis, we measured cosine similarity between RAG and Non-RAG outputs using the Gemma-3 embedding model (300M). The mean similarity was \textbf{83.54\% (SD = 8.3\%)}, suggesting that the RAG pipeline produces responses that are broadly comparable to those generated with full history access. However, we note two caveats: (1) the dense conversations were synthesised by duplicating existing message chains, which does not capture the semantic drift and topic evolution characteristic of organic long-term dialogues; and (2) embedding similarity is a proxy for semantic closeness, not a direct measure of response quality or appropriateness. A full quality evaluation with naturalistic multi-session conversations is an important direction for future work.

The latency results validate H2 with respect to scalability: the RAG pipeline maintains bounded response time independent of conversation length. The output similarity results provide preliminary evidence that this efficiency does not come at the cost of response relevance, though further validation with organic dialogue data is warranted, \textbf{supporting H2}.

\vspace{-1em}
\section{Discussion}
\label{sec:result_discussion}
Analysing Table~\ref{tab:stat_results} and Figure~\ref{fig:Agentic_vs_LLM}, ARIS consistently outperformed the LLM-only system on all four robot perception dimensions as measured by the GodSpeed questionnaire on Animacy, Intelligence, Anthropomorphism, and Likeability. Interaction order did not have a significant influence. These results demonstrate that integrating social reasoning, multimodal perception, context retrieval, and embodied action within a unified agentic architecture yields measurable improvements in user experience over a representative LLM-only baseline.

\subsection{System Performance Analysis}
To understand the system performance, we analyse targeted questionnaire items from the ARIS condition. Participants rated the robot’s use of vision at 4.17/5, with several noting its ability to comment on outfits and appearance which was a capability absent in the baseline, likely contributing to the Intelligence gain ($d = 0.74$). Engagement with robot movements was rated 3.91/5 under ARIS vs.\ 2.0/5 under LLM-only, and expected movement execution 3.86/5 vs.\ 2.10/5, suggesting the physical embodiment module is a strong driver of the Anthropomorphism ($d = 1.05$) and Animacy ($d = 0.70$) effects. Under ARIS, 7/12 second-to-interact participants were successfully recognised and 9/12 reported the robot ``knew about my partner,’’ compared to 4/12 and 4/12 under the baseline, indicating the Social World Model underpins the gains in perceived Intelligence and Anthropomorphism. While this analysis cannot establish causal isolation, it provides converging evidence that improvements are distributed across system capabilities rather than attributable to any single feature. Notably, ARIS achieves these gains cost-effectively by invoking the VLM only when vision is contextually required and capping RAG retrieval at 80 messages, reducing token consumption as dialogues grow.

\subsection{Social World Model Performance}
Incorporating a Social World Model enabled more personalized interactions, with 75\% of second-to-interact participants reporting that ARIS recognized them and the robot receiving an average rating of 4.39 out of 5 when asked, ``Did the robot seem intelligent?’’. However, there were instances where the robot failed to accurately identify participants upon introduction. These failures were traced to inaccuracies in the ASR system, which occasionally captured names incorrectly. We mitigated this using the \emph{Levenshtein distance} to match participant names with those introduced by partners (Section~\ref{sec:world-model-desc}), though recognition still failed in some cases.

\subsection{Limitations}
The present evaluation lacks a formal ablation study that systematically disables individual components (e.g., ARIS without vision, ARIS without the Social World Model); the system performance analysis above provides only preliminary evidence for each module’s contribution. Additionally, ARIS was evaluated on a single robotic platform, and the RAG scalability experiment used synthetically duplicated messages rather than organic long-term dialogue data, which does not capture semantic drift or topic evolution characteristic of real multi-session conversations. The sample size ($N = 23$) is modest, and future work should validate these findings with larger and more diverse participant pools.

\section{Future Directions}
Since the architecture is largely platform-agnostic with only client-side adaptations, future work will test across diverse platforms to assess generalisability. Key research directions include: (1) conducting formal ablation studies to isolate the contribution of individual components; (2) advancing multi-step reasoning for social contexts \cite{ahn_as_2022}; (3) enabling the system to learn from its own mistakes through adaptive feedback loops \cite{ke_hydra_2024}; and (4) evaluating the RAG pipeline with naturalistic multi-session dialogue data collected over extended deployment periods.

\section{Conclusion}
This work presented ARIS, an agentic AI framework that combines large language models, multimodal perception, and a Social World Model to improve social robotics. Our system enables robots to sustain context, reason about relationships, and coordinate dialogue with physical actions. User studies show clear gains in engagement, intelligence, and anthropomorphism compared to an LLM-only baseline. These results highlight the value of embodied reasoning and structured memory for advancing social interaction. Future work will expand modalities, improve context inference, and support long term adaptation across platforms.

\section*{Acknowledgment}
ChatGPT has been used in editing the text and generating images for Figure~\ref{fig:server_diagram} and Figure~\ref{fig:user_study_flowchart}. 
{
    \small
    \bibliographystyle{ieeenat_fullname}
    \bibliography{main}
}


\end{document}